\documentclass[letterpaper]{article} 
\usepackage{aaai2026}  
\usepackage{times}  
\usepackage{helvet}  
\usepackage{courier}  
\usepackage[hyphens]{url}  
\usepackage{graphicx} 
\usepackage{natbib}  
\usepackage{caption} 
\usepackage{algorithm}
\usepackage{algorithmic}
\usepackage{url}            
\usepackage{booktabs}       
\usepackage{amsfonts}       
\usepackage{nicefrac}       
\usepackage{microtype}      
\usepackage{xcolor}         
\usepackage{amsmath}
\usepackage{multirow}
\usepackage{colortbl}

\usepackage{newfloat}
\usepackage{listings}
\DeclareCaptionStyle{ruled}{labelfont=normalfont,labelsep=colon,strut=off} 
\floatstyle{ruled}
\newfloat{listing}{tb}{lst}{}
\floatname{listing}{Listing}
\nocopyright
\title{Commonality in Few: Few-Shot Multimodal Anomaly Detection via Hypergraph-Enhanced Memory}
\author{
    Written by AAAI Press Staff\textsuperscript{\rm 1}\thanks{With help from the AAAI Publications Committee.}\\
    AAAI Style Contributions by Pater Patel Schneider,
    Sunil Issar,\\
    J. Scott Penberthy,
    George Ferguson,
    Hans Guesgen,
    Francisco Cruz\equalcontrib,
    Marc Pujol-Gonzalez\equalcontrib
}
\author{
    Yuxuan Lin\textsuperscript{\rm 1},
    Hanjing Yan\textsuperscript{\rm 2},
    Xuan Tong\textsuperscript{\rm 3},
    Yang Chang\textsuperscript{\rm 3},
    Huanzhen Wang\textsuperscript{\rm 3},
    Ziheng Zhou\textsuperscript{\rm 3},\\
    Shuyong Gao\textsuperscript{\rm 1},
    Yan Wang\textsuperscript{\rm 4$*$},
    Wenqiang Zhang\textsuperscript{\rm 1,3}\thanks{Corresponding authors.}
}
\affiliations{
    \textsuperscript{\rm 1}Shanghai Key Lab of Intelligent Information Processing, College of Computer Science and Artificial Intelligence, Fudan University
    
    \textsuperscript{\rm 2}School of Information Science and Engineering, East China University of Science and Technology
    
    \textsuperscript{\rm 3}College of Intelligent Robotics and Advanced Manufacturing, Fudan University
    
    \textsuperscript{\rm 4}School of Data Science and Engineering, East China Normal University\\

    \{yuxuanlin24, xtong23, ychang24, hzwang24, zhouzh24\}@m.fudan.edu.cn, \{wqzhang, sy\_gao\}@fudan.edu.cn, Y21250025@mail.ecust.edu.cn, yanwang@dase.ecnu.edu.cn

}

\usepackage{bibentry}

\begin{document}

\maketitle

\begin{abstract}
Few-shot multimodal industrial anomaly detection is a critical yet underexplored task, offering the ability to quickly adapt to complex industrial scenarios. In few-shot settings, insufficient training samples often fail to cover the diverse patterns present in test samples. This challenge can be mitigated by extracting structural commonality from a small number of training samples. In this paper, we propose a novel few-shot unsupervised multimodal industrial anomaly detection method based on structural commonality, \textbf{CIF} (\textbf{C}ommonality \textbf{I}n \textbf{F}ew). To extract intra-class structural information, we employ hypergraphs, which are capable of modeling higher-order correlations, to capture the structural commonality within training samples, and use a memory bank to store this intra-class structural prior. Firstly, we design a semantic-aware hypergraph construction module tailored for single-semantic industrial images, from which we extract common structures to guide the construction of the memory bank. Secondly, we use a training-free hypergraph message passing module to update the visual features of test samples, reducing the distribution gap between test features and features in the memory bank. We further propose a hyperedge-guided memory search module, which utilizes structural information to assist the memory search process and reduce the false positive rate. Experimental results on the MVTec 3D-AD dataset and the Eyecandies dataset show that our method outperforms the state-of-the-art (SOTA) methods in few-shot settings.
\end{abstract}

\begin{links}
    \link{Code}{https://github.com/Sunny5250/CIF}
\end{links}

\section{Introduction}
\label{intro}

Anomaly detection (AD) is one of the most extensively studied tasks in computer vision, and industrial anomaly detection that this paper focuses on is a important technology for ensuring product quality and production line stability. Manual detection methods are inefficient and easily affected by human factors, making them unsuitable for large-scale, high-precision production needs. Traditional anomaly detection methods \cite{crosby1994detect,breunig2000lof,tang2002enhancing,liu2008isolation} based on data distribution struggle to detect diverse or subtle anomalies in complex industrial environments and lack robustness. Machine learning and deep learning-based methods for anomaly detection are mainly categorized into feature embedding-based methods \cite{gudovskiy2022cflow,rudolph2023asymmetric,roth2022towards,wang2023multimodal} and reconstruction-based methods \cite{zavrtanik2021reconstruction,zavrtanik2021draem,zhang2023destseg,chen2023easynet}. However, in real-world industrial environments, anomaly samples are extremely scarce and costly to obtain. As a result, most existing methods are designed in unsupervised settings. By learning the feature distribution of normal samples, unsupervised methods detect anomalies as deviations from the normal pattern.

\begin{figure}[t]
\centering
\includegraphics[width=0.9\columnwidth]{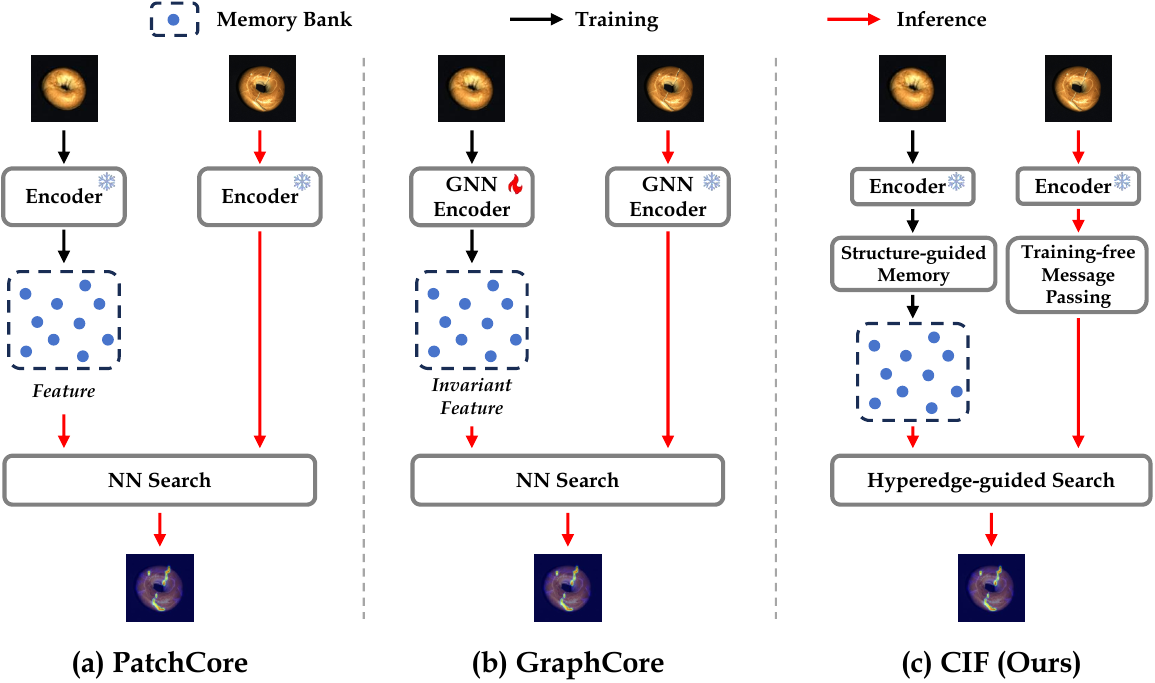}
\caption{The Main Idea of CIF. Different from (a) PatchCore and (b) GraphCore, our HyperAD (c) extracts higher-order correlations among patch features through a hypergraph and uses training-free message passing to obtain patch features enriched with structured contextual information.}
\label{motivation}
\end{figure}

Few-shot anomaly detection has been proposed to address scenarios with limited normal training samples. Compared to full-shot methods, it offers stronger transferability and faster adaptability to new environments, making it well-suited for complex and dynamic industrial settings. MetaFormer \cite{wu2021learning} uses meta-learning method, introducing additional datasets to train a general model, and then fine-tuning the model with a few normal industrial data samples. AnomalyGPT \cite{gu2024anomalygpt} introduces Large Vision-Language Models (LVLMs), leveraging their rich prior knowledge and strong generalization capabilities to perform effective few-shot anomaly detection and localization. Memory bank-based method \cite{defard2021padim,roth2022towards,lee2022cfa,wang2023multimodal,xie2023pushing} constructs a memory bank of normal visual features, which are used to compare with test sample features during testing to detect anomalies. However, in few-shot settings, the memory bank often fails to adequately cover the normal patterns present in test samples, resulting in high false positive rates. GraphCore \cite{xie2023pushing} is a memory bank-based method, using Graph Neural Networks (GNNs) to extract vision isometric invariant features in few-shot settings, which reduces redundancy in the memory bank, improving the coverage of memory bank features of the features of test samples.

For single-semantic industrial images, samples of the same class often share highly consistent structures. Graphs struggle to capture higher-order correlations between patches, whereas hypergraphs can better model these correlations and extract shared structural information among samples of the same class, enhancing memory bank coverage in few-shot settings. We propose CIF (Commonality In Few), a hypergraph-based few-shot unsupervised multimodal industrial anomaly detection method. We construct hypergraphs for industrial images and use shared structures from a few training samples to guide memory bank construction and compression. A training-free hypergraph message passing module further updates test features, narrowing the gap with memory features. During memory search, hyperedge features guide matching, reducing false positives through structural cues. As shown in Figure \ref{motivation}, PatchCore \cite{roth2022towards} detects anomalies via patch feature sampling and nearest-neighbor search, while GraphCore \cite{xie2023pushing} employs GNNs to aggregate neighborhood information and extract vision isometric invariant features. In contrast, CIF uses a pretrained encoder and training-free hypergraph message passing to model higher-order correlations, capture structural commonality, and achieve accurate anomaly detection and localization with fewer false positives.

In summary, the contributions of our work are as follows:
\begin{itemize}
\item We propose CIF, which explores the application of hypergraphs in few-shot unsupervised multimodal industrial anomaly detection, and outperforms the state-of-the-art methods on the MVTec 3D-AD and Eyecandies datasets in few-shot settings.
\item We introduce Semantic-Aware Hypergraph Construction (SAHC), which constructs a hypergraph with more evenly distributed hyperedges for single-semantic industrial images.
\item We design Structure-Guided Memory Sampling (SGMS), which uses intra-class structural commonality to guide the construction and compression of the memory bank.
\item We use Bidirectional Training-Free Hypergraph Message Passing (Bi-TF-MP) to reduce the distribution gap between test sample node features and memory bank node features.
\item We propose Hyperedge-Guided Memory Search (HGMS), which uses hyperedge features to guide the matching between test sample features and memory bank features, effectively reducing the false positive rate.
\end{itemize}

\section{Related Work}

\subsection{Industrial Anomaly Detection}

Industrial anomaly detection identifies anomaly patterns in industrial data. Traditional methods \cite{crosby1994detect,angiulli2002fast,breunig2000lof,shyu2003novel,liu2008isolation,scholkopf2001estimating} rely on statistical or distance-based analysis. With advances in deep learning and computer vision, current deep learning-based methods are mainly unsupervised, covering both single-modal and multimodal settings (e.g., RGB images, 3D point clouds). They are generally categorized into feature embedding-based and reconstruction-based methods. Feature embedding-based methods use pretrained extractors to obtain features for detection and segmentation, including teacher-student architecture methods \cite{bergmann2020uninformed,zhang2023destseg,rudolph2023asymmetric}, one-class classification methods \cite{li2021cutpaste,liu2023simplenet}, distribution map methods \cite{yu2021fastflow,gudovskiy2022cflow}, and memory bank methods \cite{roth2022towards,wang2023multimodal,xie2023pushing}. Reconstruction-based methods \cite{zavrtanik2021draem,zhang2024realnet,chen2023easynet} reconstruct test samples to normal patterns and detect anomalies via reconstruction differences. Few-shot anomaly detection learns compact, generalizable features from limited data, offering better flexibility and transferability than full-shot approaches. Meta learning-based methods \cite{wu2021learning,huang2022registration} train generalizable models adaptable to new domains with minimal fine-tuning. Recently, large model-based methods \cite{gu2024anomalygpt,lee2024text,li2024promptad,zhu2024toward} leverage the rich priors and strong perception abilities of large models, achieving powerful few-shot and transfer performance.

\begin{figure*}[t]
\centering
\includegraphics[width=0.9\textwidth]{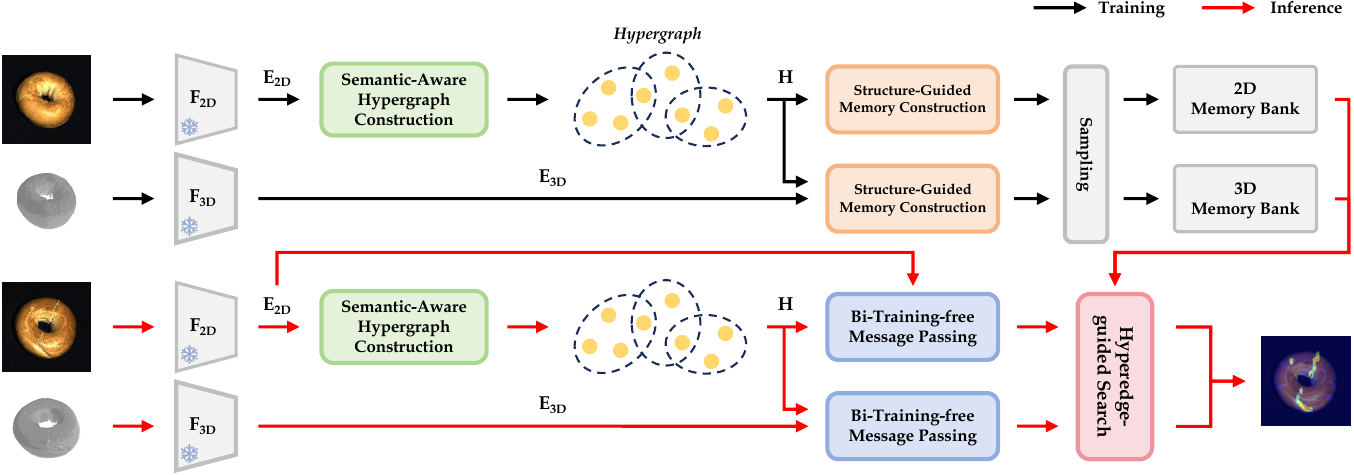}
\caption{\textbf{The pipeline of CIF}. Our CIF contains four important parts: (1) Semantic-Aware Hypergraph Construction (SAHC), which constructs hypergraphs for single-semantic samples based on clustering. (2) Structure-Guided Memory Sampling (SGMS), which uses structural commonality in the hypergraph to guide the construction and compression of the memory bank. (3) Bidirectional Training-Free Hypergraph Message Passing (Bi-TF-MP), which performs bidirectional message passing between test samples and the memory bank to reduce their distribution gap. (4) Hyperedge-Guided Memory Search (HGMS), which uses a hyperedge-guided two-stage search to reduce the false positive rate of detection.}
\label{pipeline}
\end{figure*}

\subsection{Hypergraph Learning}

Hypergraphs are a generalized form of graphs that allow an edge (hyperedge) to connect multiple nodes, effectively representing high-order correlations and complex structures. Zhou et al. \cite{zhou2006learning} first introduces hypergraph learning to minimize the label difference among correlational nodes. HGNN \cite{feng2019hypergraph} and its extended version HGNN+ \cite{gao2022hgnn+} designes message passing frameworks between hyperedges and nodes, constructing more general hypergraph neural networks to learn representations of high-order data structures. ViHGNN \cite{han2023vision} applies hypergraph neural networks to computer vision, employing fuzzy clustering algorithm (fuzzy c-means) to construct hyperedges among image patches and capture high-order correlations, achieving high accuracy in tasks such as image classification and object detection. Based on ViHGNN, DVHGNN \cite{li2025dvhgnn} introduces dilated hyperedges to capture multi-scale sparse dependencies. While these hypergraph learning methods are generalizable across various tasks, they require adjustments to adapt to industrial data. Therefore, using hypergraphs for industrial anomaly detection remains a challenge.

\section{Method}

\subsection{Preliminary: Hypergraph Definition}

In traditional graph theory, a graph $G = (V, E)$ represents binary correlations between entities, where $V$ is the set of nodes and $E \subseteq V \times V$ is the set of edges connecting pairs of nodes. However, such binary correlations often struggle to capture the complex higher-order interactions among multiple entities. To address this issue, a hypergraph provides a structure capable of expressing higher-order correlations. As a generalization of a graph, a hypergraph is defined as $\mathcal{G} = (V, \mathcal{E})$, where $V$ is the set of nodes (vertices) and $\mathcal{E}$ is the set of hyperedges. Each hyperedge $e \in \mathcal{E}$ is a non-empty subset of nodes, i.e., $e \subseteq V$, with $|e| \geq 2$. Unlike an ordinary graph edge that connects exactly two nodes, a hyperedge can simultaneously connect an arbitrary number of nodes, enabling the modeling of higher-order correlations among multiple entities. A hypergraph is commonly represented by an incidence matrix $\mathbf{H} \in \mathbb{R}^{|V| \times |\mathcal{E}|}$, where $\mathbf{H}(v, e) = 1$ if vertex $v$ belongs to hyperedge $e$, and $0$ otherwise. Based on incidence matrix, various hypergraph Laplacian operators can be constructed to support spectral analysis and learning tasks.

\subsection{Overview}

Our CIF is a memory bank-based anomaly detection method that enhances memory construction and search through hypergraph-based structural representations. Unlike other memory bank-based methods, we use intra-class structural priors from training samples to guide the construction and compression of the memory bank. Moreover, we do not directly perform nearest-neighbor search to match the features of test samples and those in the memory bank. Instead, we use hyperedge features to match the structural information of the test samples with that in the memory bank, and then perform nearest-neighbor search guided by this structural alignment. To ensure structural consistency between memory bank and test samples in few-shot settings, we apply bidirectional training-free hypergraph message passing to update the features of test samples, thereby narrowing the distribution gap between them and the memory bank features. We use two pretrained feature extractors: DINO \cite{caron2021emerging} for extracting 2D image features, and PointMAE \cite{pang2022masked} for extracting 3D point cloud features.

\begin{figure*}[t]
\centering
\includegraphics[width=0.8\textwidth]{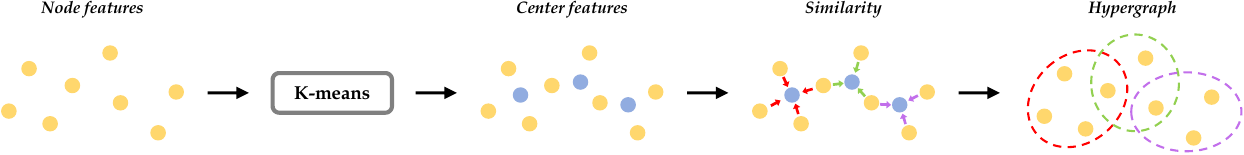}
\caption{The pipeline of Hypergraph construction. We use a clustering algorithm to compute the cluster centers among node features, which serve as hyperedge centers. We then calculate the similarity between each hyperedge center and all foreground node features, and determine the hyperedge membership of each node using a predefined threshold, resulting in the hypergraph incidence matrix.}
\label{hypergraph}
\end{figure*}

\subsection{Semantic-Aware Hypergraph Construction}
\label{hc}

Traditional hard clustering algorithms (e.g., K-Means) cannot assign nodes to multiple hyperedges, while fuzzy clustering (e.g., Fuzzy C-Means \cite{bezdek1984fcm}) results in uneven hyperedges when applied to single-semantic industrial images. We give a comparison of different hypergraph construction methods in Appendix. We propose a \textbf{S}emantic-\textbf{A}ware \textbf{H}ypergraph \textbf{C}onstruction (\textbf{SAHC}) method, which first generates a set of centers (hyperedges) using K-Means. Then, we calculate the similarity between patch (node) features and the centers, and determine the assignment of nodes to hyperedges based on a predefined threshold.

For image data, we use a pretrained feature extractor to obtain patch features $X = [x_1, x_2, ..., x_N]$, where each patch feature can be viewed as a node in the set $V = \{v_1, v_2, ..., v_N\}$. To reduce the interference of background information during hypergraph construction, we extract the foreground mask of each sample using its 3D point cloud, and filter out the foreground nodes from $V$ to obtain the foreground node set $V_{\text{fore}}$. As shown in Figure \ref{hypergraph}, we then perform K-means clustering on $V_{\text{fore}}$ to obtain $|\mathcal{E}|$ cluster centers, which serve as hyperedge centers, forming the hyperedge set $\mathcal{E}$. We calculate the cosine similarity between all node features in $V_{\text{fore}}$ and the hyperedge centers, and apply min-max normalization. Using the similarity matrix, we determine the assignment of each node to one or more hyperedges based on a predefined threshold, resulting in a incidence matrix $\mathbf{H} \in \mathbb{R}^{|V| \times |\mathcal{E}|}$. Due to the needs of subsequent modules, we also compute a hard incidence matrix $\mathbf{H}_{\text{hard}} \in \mathbb{R}^{|V| \times 1}$, where each node is assigned only to one hyperedge with the highest similarity.

During the hypergraph construction process, we observed that hypergraphs built from RGB image features of industrial samples are more balanced and reliable, while those constructed from 3D point cloud features exhibit significant imbalance. Therefore, we uniformly use hypergraphs constructed from RGB image features to support the subsequent modules in both the 2D and 3D modals.

\subsection{Structure-Guided Memory Sampling}
\label{ms}

\begin{figure}[t]
\centering
\includegraphics[width=0.8\columnwidth]{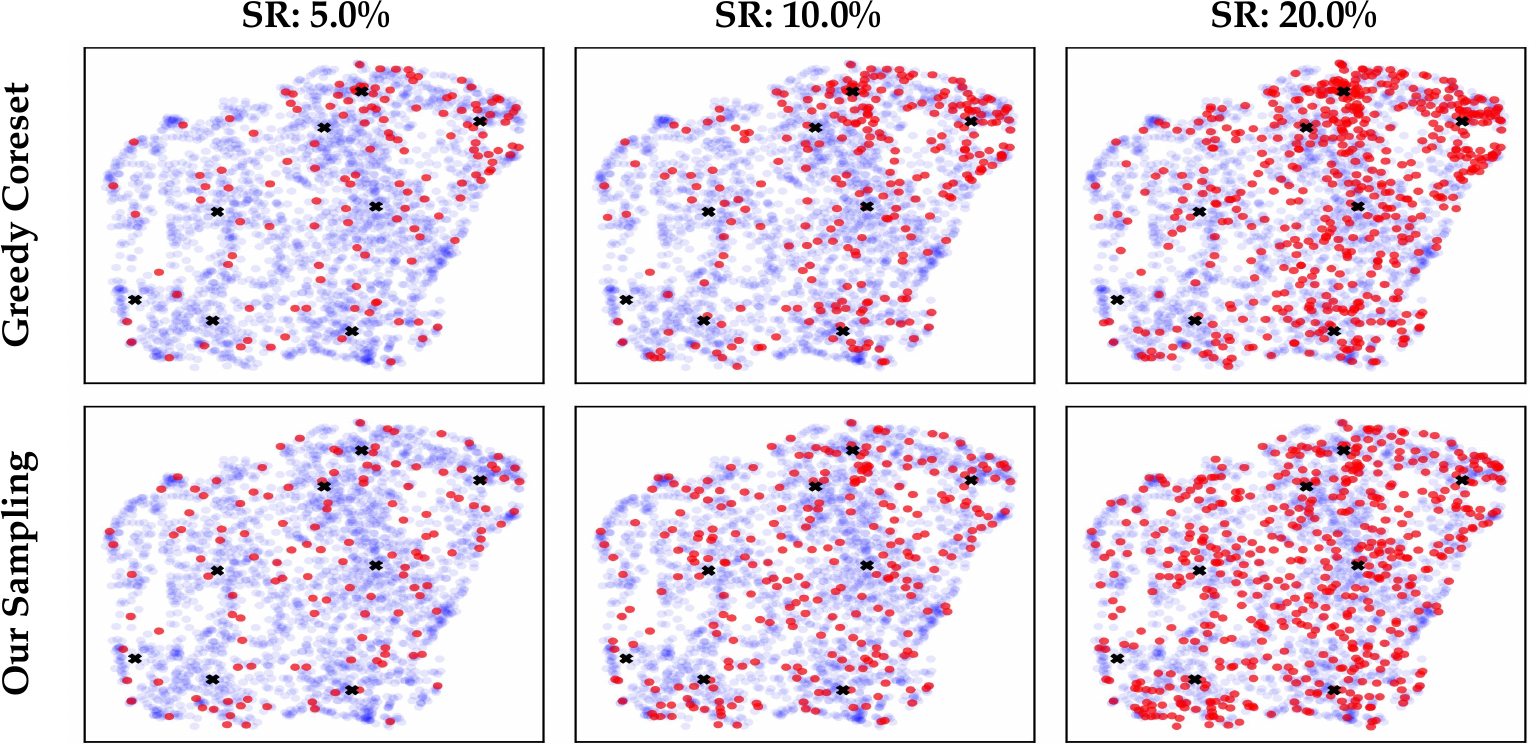}
\caption{Comparison: Greedy Coreset Sampling (Top) vs. Our Sampling (Bottom) at different sampling rates (SR). Red points represent features before sampling, blue points represent features after sampling, and black crosses represent hyperedge features.}
\label{hypergraph_sampling}
\end{figure}

For single-semantic industrial images, samples of the same class often exhibit consistent structural patterns. Such intra-class structural priors can be used to guide the matching between memory bank features and test sample features, and can alleviate the issue of sparse information in the memory bank. Using SAHC, we obtain the incidence matrices and hyperedge features of training samples, and then design a \textbf{S}tructure-\textbf{G}uided \textbf{M}emory \textbf{S}ampling (\textbf{SGMS}) module to update and compress the memory bank. We fix the number of hyperedges per class, and use the patch features, incidence matrix, and hyperedge features of the first training sample to initialize the memory bank for this class. The memory bank update process consists of three stages: node assignment, hyperedge update, and memory bank sampling.

In the node assignment stage, we first compute the average patch features for each hyperedge in the memory bank to obtain the corresponding hyperedge features. For a newly input training sample, we compute its hyperedge features and calculate the pairwise distances between these and the hyperedge features in the memory bank. To facilitate node assignment, we use $\mathbf{H}_{\text{hard}}$ as the reference for node-hyperedge membership. Based on this, we merge the nodes from each hyperedge in the training sample with the most similar hyperedge in the memory bank. In the hyperedge update stage, after assigning the node features of a new sample , we update the hyperedge features of the memory bank by computing the merged hyperedge features, preparing them for the next node assignment step. In the memory bank sampling stage, existing methods \cite{roth2022towards,xie2023pushing} perform greedy coreset sampling across all nodes in the memory bank. In contrast, we perform greedy coreset sampling within each hyperedge individually. If a hyperedge contains fewer than one node after sampling, we select a single node from this hyperedge such that the maximum distance from this node to all other nodes in the hyperedge is minimized. Figure \ref{hypergraph_sampling} visually compares our sampling method with greedy coreset sampling.

\subsection{Bidirectional Training-Free Hypergraph Message Passing}
\label{mp}

To mitigate the distribution gap between test sample node features and memory bank node features caused by the limited number of features in the memory bank, we leverage intra-class structural priors from the memory bank and the structural information of the test sample. We propose a \textbf{Bi}directional \textbf{T}raining-\textbf{F}ree hypergraph \textbf{M}essage \textbf{P}assing (\textbf{Bi-TF-MP}) module, which constructs a joint hypergraph between test sample node features and memory bank node features to explicitly connect them. We then perform $L$ layers of training-free hypergraph message passing on this joint hypergraph, enabling information exchange between each node and its $L$-hop neighbors in one step.

We denote the test hypergraph as $\mathcal{G}^{test} = (X^{test} \in \mathbb{R}^{N \times dim}, \mathbf{H}^{test} \in \{0,1\}^{N \times \mathcal{E}_{num}})$ and the memory bank hypergraph as $\mathcal{G}^{mem} = (X^{mem} \in \mathbb{R}^{A \times dim}, \mathbf{H}^{mem} \in \{0,1\}^{A \times \mathcal{E}_{num}})$, where $dim$ is the feature dimension of each node, $\mathbf{H}^{*}$ is the incidence matrix, $N$ is the number of nodes in the test sample, and $A$ is the number of nodes in the memory bank.

We compute the cosine similarity between $X^{test}$ and $X^{mem}$, and for each test node $x^{test}_i$, we select the top-$k$ most similar memory bank nodes. We construct a hyperedge that includes $x^{test}_i$ and the selected $k$ memory bank nodes. In total, $N$ such cross-domain hyperedges are constructed, resulting in an incidence matrix $\mathbf{H}^{t \rightarrow m} \in \{0,1\}^{(A+N) \times N}$. Similarly, for each memory node $x^{mem}_i$, we select the top-$k$ most similar test nodes, and construct a hyperedge that includes $x^{mem}_i$ and the selected $k$ test nodes. In total, $A$ such cross-domain hyperedges are constructed, resulting in an incidence matrix $\mathbf{H}^{m \rightarrow t} \in \{0,1\}^{(A+N) \times A}$. By concatenating $\mathbf{H}^{t \rightarrow m}$ and $\mathbf{H}^{m \rightarrow t}$ along the hyperedge dimension, we obtain the cross-domain incidence matrix $\mathbf{H}^{cross} \in \{0,1\}^{(A+N) \times (A+N)}$.

To enable message passing across all three domains (test sample domain, memory bank domain, and cross-domain), we apply zero-padding to extend $\mathbf{H}^{test} \in \{0,1\}^{N \times \mathcal{E}_{num}}$ to $\widetilde{\mathbf{H}}^{test} \in \{0,1\}^{(N+A) \times \mathcal{E}_{num}}$, and similarly extend $\mathbf{H}^{mem} \in \{0,1\}^{A \times \mathcal{E}_{num}}$ to $\widetilde{\mathbf{H}}^{mem} \in \{0,1\}^{(N+A) \times \mathcal{E}_{num}}$. We then concatenate $\widetilde{\mathbf{H}}^{test}$, $\widetilde{\mathbf{H}}^{mem}$, and $\mathbf{H}^{cross}$ along the hyperedge dimension to obtain the final joint incidence matrix $\mathbf{H}^{joint}$.

\begin{equation}
\begin{split}
    \mathbf{H}^{joint} &= \bigl[ \underbrace {\widetilde{ \mathbf{H}}^{test}}_{test} \big| \underbrace{ \widetilde{ \mathbf{H}}^{mem}}_{memory} \big| \underbrace{ \mathbf{H}^{cross}}_{cross} \bigr] \\
    &\in \{0,1\}^{(N+A) \times \bigl( 2\mathcal{E}_{num}+N+A\bigr)}
\end{split}
\end{equation}

\noindent then we concatenate the test sample node features $X^{test}$ and the memory bank node features $X^{mem}$ to obtain the joint node features $X^{joint}$:

\begin{equation}
    X^{joint} = \bigl[ \underbrace {X^{test}}_{test} \big| \underbrace{X^{mem}}_{memory} \bigr] \in \mathbb{R}^{(N+A) \times dim}
\end{equation}

We follow TF-MP \cite{tang2024training} to perform message passing on the joint hypergraph. For a hypergraph $\mathcal{G} = (X, \mathbf{H})$, we first compute the transition matrix $\mathbf{A}_{trans}$, and then compute the training-free message passing kernel $\mathbf{S}$ as follows:

\begin{equation}
\begin{split}
&\mathbf{S}=(1-\alpha)^{L}\mathbf{A}_{trans}^{L}+\alpha\sum_{l=0}^{L-1}(1-\alpha)^{l}\mathbf{A}_{trans}^{l}
,\\
&\mathbf{A}_{trans}=\mathbf{D}_{v}^{-\frac12}\left( \mathbf{H}\mathbf{D}_{e}^{-1}\mathbf{H}^{\top}+\mathbf{I}_{N}\right)\mathbf{D}_{v}^{-\frac12}
\end{split}
\end{equation}

\noindent where $\mathbf{D}_{e}$ is the hyperedge degree matrix of $\mathbf{H}$, $\mathbf{D}_{v}$ is the node degree matrix of $\widetilde{\mathbf{W}}=\mathbf{H}\mathbf{D}_{e}^{-1}\mathbf{H}^{\top}+\mathbf{I}_{N}$, and $\mathbf{I}_{N}$ is the identity matrix. Moreover, $\alpha$ is the retention coefficient, a larger $\alpha$ emphasizes the preservation of the own information of a node, while a smaller $\alpha$ increases the influence of neighboring nodes. $L$ denotes the number of message passing layers, allowing any node to exchange information with its $L$-hop neighbors in the hypergraph.

We apply the training-free message passing kernel $\mathbf{S}$ to the joint hypergraph $\mathcal{G}^{joint} = (X^{joint} \in \mathbb{R}^{(N+A) \times dim}, \mathbf{H}^{joint} \in \{0,1\}^{(N+A) \times \bigl( 2\mathcal{E}_{num}+N+A\bigr)})$ to obtain the updated node features of test samples, denoted as $X^{test}_{new}$:

\begin{equation}
\begin{split}
&X^{joint}_{new} = (\mathbf{S}X^{joint})^{\top}\in \mathbb{R}^{(N+A)\times dim} \;\Longrightarrow \\
&X^{test}_{new}=X^{joint}_{new\,[:,\,:N]}\in\mathbb{R}^{N\times dim}
\end{split}
\end{equation}

\subsection{Hyperedge-Guided Memory Search}

To match structurally consistent test sample nodes and memory bank nodes based on intra-class structural priors, we design a \textbf{H}yperedge-\textbf{G}uided \textbf{M}emory \textbf{S}earch (\textbf{HGMS}) module that introduces hyperedge features to assist in matching test sample features with those in the memory bank, enabling more targeted search based on intra-class structural priors. Specifically, we compute pairwise cosine similarity between the updated hyperedge features (obtained through Bi-TF-MP) of the test sample and the hyperedge features in the memory bank. For the $i$-th hyperedge in the test sample, we collect all nodes within this hyperedge to form a test node subset $X^{test}_{sub\: i}$. We then select the top-$k$ most similar hyperedges from the memory bank and collect all memory nodes within these $k$ hyperedges to form a memory bank node subset $\mathcal{M}_{sub\: i}$. A patch-level nearest neighbor search is then performed between $X^{test}_{sub\: i}$ and $\mathcal{M}_{sub\: i}$ to compute the patch-level anomaly score $\mathcal{A}$ for the test sample:

\begin{equation}
\mathcal{A}_{ij} = \min_{m \in \mathcal{M}_{sub\: i}} \| X^{test}_{ij} - m \|_2
\end{equation}

\noindent where $\mathcal{A}_{ij}$ and $X^{test}_{ij}$ denote the anomaly score and patch feature for the $j$-th patch within the $i$-th hyperedge of the test sample, respectively. $m$ represents a node feature from the memory bank node subset $\mathcal{M}_{sub: i}$ corresponding to the $i$-th hyperedge in the test sample. Additionally, we retain the conventional patch-level search method that directly compares test sample node features with memory bank node features. The resulting scores are element-wise multiplied with $\mathcal{A}$ to obtain the final anomaly scores.

\begin{table*}
  \centering
  \caption{I‑AUROC/AUPRO scores for anomaly detection and localization of all categories of MVTec 3D‑AD and Eyecandies in few‑shot settings (1‑shot, 2‑shot and 4‑shot). The best is in \textcolor{red}{red} and the second best is in \textcolor{blue}{blue}. TB and TF represent Training-Based and Training-Free, respectively.}
  \renewcommand\arraystretch{0.8}
  \resizebox{1.0\linewidth}{!}{
  \begin{tabular}{cll|cccccccccc|c}
  
    \toprule
    \multicolumn{14}{c}{\textbf{MVTec 3D-AD}} \\
    \toprule
    
    Setting & Type & Method & Bagel & Cable Gland & Carrot & Cookie & Dowel & Foam & Peach & Potato & Rope & Tire & Mean \\
    \midrule
    \multirow{8}*{\rotatebox{90}{1-shot}} & \multirow{5}*{TB}
    & AST \cite{rudolph2023asymmetric} & 70.7/75.9 & 42.2/73.3 & 54.8/88.0 & 49.0/60.2 & 53.8/79.4 & 46.4/44.0 & 51.9/84.0 & 49.7/85.9 & 72.0/75.8 & 41.9/74.0 & 53.2/74.0 \\
    & & EasyNet \cite{chen2023easynet} & 61.4/79.6 & 21.2/75.1 & 52.0/91.0 & 75.9/69.8 & 56.5/\textcolor{blue}{85.8} & 62.8/49.4 & 65.7/69.0 & 63.0/88.1 & \textcolor{red}{94.6}/71.8 & 47.7/75.4 & 60.1/75.5 \\
    & & ShapeGuided \cite{chu2023shape} & 65.9/\textcolor{red}{95.9} & 44.4/71.6 & 62.3/93.5 & \textcolor{blue}{93.8}/\textcolor{red}{94.1} & 59.3/\textcolor{red}{86.4} & 57.6/63.8 & 67.6/94.0 & 42.8/\textcolor{red}{96.3} & \textcolor{blue}{93.3}/88.8 & \textcolor{red}{62.9}/\textcolor{blue}{90.1} & 65.0/87.4 \\
    & & M3DM \cite{wang2023multimodal} & \textcolor{red}{87.8}/\textcolor{blue}{95.3} & \textcolor{blue}{64.1}/\textcolor{red}{81.5} & \textcolor{red}{78.0}/\textcolor{red}{97.2} & 92.7/90.3 & \textcolor{red}{64.2}/81.6 & 65.3/\textcolor{blue}{82.0} & \textcolor{blue}{75.5}/94.0 & \textcolor{red}{79.8}/94.8 & 85.8/\textcolor{blue}{95.2} & 45.4/89.8 & \textcolor{red}{73.9}/\textcolor{blue}{90.2} \\
    & & CFM \cite{costanzino2024multimodal} & 43.5/92.6 & 56.7/79.4 & \textcolor{blue}{76.4}/\textcolor{blue}{96.9} & \textcolor{red}{95.4}/\textcolor{blue}{92.8} & 53.2/85.2 & \textcolor{blue}{71.4}/\textcolor{red}{87.8} & 63.0/\textcolor{blue}{94.9} & \textcolor{blue}{64.2}/\textcolor{blue}{96.1} & 91.8/\textcolor{red}{96.6} & \textcolor{blue}{59.2}/\textcolor{red}{91.4} & 67.5/\textcolor{red}{91.4} \\
    \cmidrule(lr{0pt}){2-14}
    & \multirow{2}*{TF} & Patchcore+FPFH \cite{horwitz2023back} & 62.2/92.8 & 53.4/76.8 & 54.0/96.7 & 55.9/\textcolor{blue}{92.8} & 54.7/84.6 & 63.3/71.9 & 49.6/\textcolor{red}{95.9} & 60.7/\textcolor{blue}{96.1} & 88.8/90.8 & 56.6/84.1 & 59.9/88.3 \\
    & & \cellcolor{gray!20}CIF (Ours) & \cellcolor{gray!20}\textcolor{blue}{78.9}/85.2 &  \cellcolor{gray!20}\textcolor{red}{68.7}/\textcolor{blue}{79.5} & \cellcolor{gray!20}72.2/95.6 & \cellcolor{gray!20}81.2/86.2 & \cellcolor{gray!20}\textcolor{blue}{62.9}/82.0 & \cellcolor{gray!20}\textcolor{red}{72.8}/70.3 & \cellcolor{gray!20}\textcolor{red}{83.9}/93.3 & \cellcolor{gray!20}60.4/94.2 & \cellcolor{gray!20}83.5/88.1 & \cellcolor{gray!20}55.8/86.4 & \cellcolor{gray!20}\textcolor{blue}{72.0}/86.1 \\
    
    \midrule
    
    \multirow{8}*{\rotatebox{90}{2-shot}} & \multirow{5}*{TB}
    & AST \cite{rudolph2023asymmetric} & 71.9/75.9 & 43.4/74.0 & 54.5/87.8 & 50.8/62.2 & 53.7/79.5 & 46.1/43.6 & 51.6/83.7 & 50.4/85.6 & 75.8/76.5 & 40.2/72.8 & 53.8/74.2 \\
    & & EasyNet \cite{chen2023easynet} & 47.6/77.8 & \textcolor{red}{76.1}/62.9 & 52.6/92.6 & 60.2/59.1 & 31.7/58.8 & 52.3/57.2 & 71.9/21.1 & \textcolor{red}{76.1}/15.2 & 61.2/43.1 & 51.2/4.7 & 58.1/49.3 \\
    & & ShapeGuided \cite{chu2023shape} & 47.9/\textcolor{red}{96.6} & 46.0/73.2 & 60.5/96.5 & \textcolor{blue}{95.9}/\textcolor{red}{95.5} & 55.3/\textcolor{blue}{86.5} & 50.2/71.4 & 69.7/95.3 & 41.3/\textcolor{blue}{96.3} & \textcolor{red}{93.6}/89.3 & \textcolor{red}{79.3}/\textcolor{blue}{91.3} & 64.0/89.2 \\
    & & M3DM \cite{wang2023multimodal} & \textcolor{red}{91.8}/\textcolor{blue}{95.5} & 57.0/\textcolor{red}{82.9} & \textcolor{blue}{79.8}/\textcolor{blue}{97.2} & 94.5/88.0 & \textcolor{blue}{61.4}/\textcolor{red}{87.0} & \textcolor{red}{79.5}/\textcolor{blue}{79.6} & \textcolor{red}{79.2}/95.1 & \textcolor{blue}{75.1}/94.2 & \textcolor{blue}{92.8}/\textcolor{blue}{95.5} & 54.1/91.0 & \textcolor{red}{76.5}/\textcolor{blue}{90.6} \\
    & & CFM \cite{costanzino2024multimodal} & 82.6/95.3 & \textcolor{blue}{65.0}/\textcolor{blue}{80.3} & \textcolor{red}{80.3}/\textcolor{red}{97.9} & \textcolor{red}{97.4}/93.1 & 53.3/85.6 & \textcolor{blue}{68.0}/\textcolor{red}{89.1} & 70.4/\textcolor{blue}{96.1} & 74.7/\textcolor{blue}{96.3} & 92.1/\textcolor{red}{96.9} & 64.0/\textcolor{red}{94.1} & \textcolor{blue}{74.8}/\textcolor{red}{92.5} \\
    \cmidrule(lr{0pt}){2-14}
    & \multirow{2}*{TF} & Patchcore+FPFH \cite{horwitz2023back} & 63.2/94.6 & 47.7/76.5 & 55.4/96.7 & 64.5/\textcolor{blue}{93.4} & 58.3/85.0 & 61.1/67.4 & 55.0/\textcolor{red}{96.2} & 56.6/\textcolor{red}{96.6} & 88.2/91.0 & \textcolor{blue}{64.4}/88.3 & 61.4/88.6 \\
    & & \cellcolor{gray!20}CIF (Ours) & \cellcolor{gray!20}\textcolor{blue}{85.3}/87.1 &  \cellcolor{gray!20}62.9/79.5 & \cellcolor{gray!20}74.0/95.8 & \cellcolor{gray!20}72.2/87.1 & \cellcolor{gray!20}\textcolor{red}{64.6}/82.1 & \cellcolor{gray!20}\textcolor{red}{79.5}/75.3 & \cellcolor{gray!20}\textcolor{blue}{77.6}/93.7 & \cellcolor{gray!20}66.9/94.9 & \cellcolor{gray!20}86.1/88.7 & \cellcolor{gray!20}62.7/87.7 & \cellcolor{gray!20}73.2/87.2 \\
    
    \midrule
    
    \multirow{8}*{\rotatebox{90}{4-shot}} & \multirow{5}*{TB}
    & AST \cite{rudolph2023asymmetric} & 70.1/74.7 & 42.9/73.7 & 55.7/87.7 & 51.8/61.3 & 54.0/79.6 & 46.6/41.3 & 52.0/84.3 & 49.8/85.9 & 72.6/75.9 & 39.8/74.2 & 53.5/73.9 \\
    & & EasyNet \cite{chen2023easynet} & 67.5/70.7 & 36.3/13.8 & 54.7/86.9 & 69.1/72.0 & \textcolor{red}{72.4}/39.3 & 50.2/53.6 & 74.3/86.2 & 63.1/90.2 & 44.9/15.5 & 53.7/66.3 & 58.6/56.5 \\
    & & ShapeGuided \cite{chu2023shape} & 65.4/\textcolor{red}{97.3} & 48.8/78.9 & 73.1/\textcolor{blue}{97.3} & \textcolor{blue}{96.5}/\textcolor{red}{95.4} & 69.8/\textcolor{red}{90.4} & 59.1/83.6 & 68.1/95.7 & 49.9/\textcolor{red}{97.5} & \textcolor{blue}{92.2}/89.6 & \textcolor{red}{75.1}/92.1 & 69.8/91.8 \\
    & & M3DM \cite{wang2023multimodal} & \textcolor{red}{98.6}/96.1 & \textcolor{blue}{68.5}/\textcolor{red}{87.2} & \textcolor{blue}{83.7}/\textcolor{blue}{97.3} & 93.8/90.5 & 59.6/86.5 & \textcolor{red}{87.8}/\textcolor{blue}{86.6} & \textcolor{blue}{85.3}/96.4 & 70.5/94.6 & 89.3/\textcolor{blue}{95.5} & 56.1/\textcolor{blue}{93.1} & \textcolor{blue}{79.3}/\textcolor{blue}{92.4} \\
    & & CFM \cite{costanzino2024multimodal} & \textcolor{blue}{92.8}/\textcolor{blue}{96.5} & 64.3/\textcolor{blue}{84.4} & \textcolor{red}{87.8}/\textcolor{red}{98.0} & \textcolor{red}{98.4}/93.5 & 64.1/\textcolor{blue}{89.7} & 77.8/\textcolor{red}{92.3} & 85.2/\textcolor{red}{97.2} & \textcolor{red}{75.3}/96.8 & \textcolor{red}{93.3}/\textcolor{red}{97.1} & 61.5/\textcolor{red}{94.8} & \textcolor{red}{80.1}/\textcolor{red}{94.0} \\
    \cmidrule(lr{0pt}){2-14}
    & \multirow{2}*{TF} & Patchcore+FPFH \cite{horwitz2023back} & 52.3/95.7 & 54.0/79.7 & 59.2/97.2 & 62.9/\textcolor{blue}{95.1} & 59.4/87.5 & 58.2/75.3 & 61.6/\textcolor{blue}{96.5} & 70.5/\textcolor{blue}{97.3} & 91.8/91.2 & \textcolor{blue}{73.5}/88.0 & 64.3/90.4 \\
    & & \cellcolor{gray!20}CIF (Ours) & \cellcolor{gray!20}91.8/92.9 &  \cellcolor{gray!20}\textcolor{red}{70.0}/83.3 & \cellcolor{gray!20}77.5/96.9 & \cellcolor{gray!20}83.4/86.4 & \cellcolor{gray!20}\textcolor{blue}{70.2}/84.7 & \cellcolor{gray!20}\textcolor{blue}{79.4}/84.1 & \cellcolor{gray!20}\textcolor{red}{85.6}/95.2 & \cellcolor{gray!20}\textcolor{blue}{75.0}/95.5 & \cellcolor{gray!20}89.8/89.3 & \cellcolor{gray!20}53.4/88.2 & \cellcolor{gray!20}77.6/89.6 \\
    
    \toprule
    \multicolumn{14}{c}{\textbf{Eyecandies}} \\
    \toprule
    
    Setting & Type & Method & Can. C. & Cho. C. & Cho. P. & Conf. & Gum. B. & Haz. T. & Lic. S. & Lollip. & Marsh. & Pep. C. & Mean \\
    \midrule
    \multirow{4}*{\rotatebox{90}{1-shot}} & \multirow{2}*{TB}
    & M3DM \cite{wang2023multimodal} & 36.2/86.8 & \textcolor{blue}{66.9}/\textcolor{red}{82.5} & \textcolor{red}{73.1}/\textcolor{red}{70.6} & \textcolor{blue}{84.8}/\textcolor{red}{94.2} & \textcolor{red}{71.3}/\textcolor{blue}{74.9} & 50.2/\textcolor{blue}{54.8} & \textcolor{blue}{57.4}/\textcolor{red}{71.0} & \textcolor{blue}{59.9}/\textcolor{blue}{84.2} & 60.3/\textcolor{red}{89.8} & \textcolor{blue}{81.0}/\textcolor{red}{90.0} & \textcolor{blue}{64.1}/\textcolor{red}{79.9} \\
    & & CFM \cite{costanzino2024multimodal} & \textcolor{red}{40.8}/\textcolor{red}{91.4} & 48.8/\textcolor{blue}{81.5} & \textcolor{red}{73.1}/\textcolor{blue}{69.7} & \textcolor{red}{88.8}/\textcolor{blue}{91.9} & 55.1/\textcolor{red}{77.4} & \textcolor{red}{72.0}/\textcolor{red}{66.1} & 45.4/\textcolor{blue}{63.6} & 59.4/82.9 & \textcolor{blue}{76.5}/\textcolor{blue}{85.1} & 57.4/\textcolor{blue}{80.9} & 61.7/\textcolor{blue}{79.1} \\
    \cmidrule(lr{0pt}){2-14}
    & TF & \cellcolor{gray!20}CIF (Ours) & \cellcolor{gray!20}\textcolor{blue}{38.1}/\textcolor{blue}{88.4} &  \cellcolor{gray!20}\textcolor{red}{81.8}/71.4 & \cellcolor{gray!20}\textcolor{blue}{70.2}/56.3 & \cellcolor{gray!20}84.0/86.9 & \cellcolor{gray!20}\textcolor{blue}{59.6}/63.9 & \cellcolor{gray!20}\textcolor{blue}{59.7}/50.6 & \cellcolor{gray!20}\textcolor{red}{58.1}/55.7 & \cellcolor{gray!20}\textcolor{red}{68.6}/\textcolor{red}{85.8} & \cellcolor{gray!20}\textcolor{red}{89.1}/65.7 & \cellcolor{gray!20}\textcolor{red}{85.4}/67.1 & \cellcolor{gray!20}\textcolor{red}{69.5}/69.2 \\
    
    \midrule
    
    \multirow{4}*{\rotatebox{90}{2-shot}} & \multirow{2}*{TB}
    & M3DM \cite{wang2023multimodal} & \textcolor{blue}{38.9}/81.1 & \textcolor{blue}{67.5}/\textcolor{red}{84.4} & \textcolor{red}{81.1}/\textcolor{blue}{68.6} & \textcolor{blue}{92.3}/\textcolor{red}{97.3} & \textcolor{blue}{61.7}/\textcolor{blue}{76.1} & 53.6/\textcolor{blue}{57.6} & \textcolor{red}{59.4}/\textcolor{red}{72.8} & 64.0/\textcolor{blue}{85.6} & 76.5/\textcolor{red}{89.3} & \textcolor{blue}{86.9}/\textcolor{red}{90.8} & \textcolor{blue}{68.2}/\textcolor{blue}{80.4} \\
    & & CFM \cite{costanzino2024multimodal} & 38.2/\textcolor{red}{92.4} & 63.2/\textcolor{blue}{84.3} & 76.5/\textcolor{red}{73.2} & 87.4/\textcolor{blue}{92.5} & 61.1/\textcolor{red}{79.5} & \textcolor{red}{72.5}/\textcolor{red}{70.3} & 51.8/\textcolor{blue}{64.9} & \textcolor{blue}{64.9}/83.2 & \textcolor{blue}{79.4}/\textcolor{blue}{86.3} & 67.8/\textcolor{blue}{81.5} & 66.3/\textcolor{red}{80.8} \\
    \cmidrule(lr{0pt}){2-14}
    & TF & \cellcolor{gray!20}CIF (Ours) & \cellcolor{gray!20}\textcolor{red}{41.8}/\textcolor{blue}{87.0} &  \cellcolor{gray!20}\textcolor{red}{79.0}/76.9 & \cellcolor{gray!20}\textcolor{blue}{78.6}/57.9 & \cellcolor{gray!20}\textcolor{red}{94.1}/85.5 & \cellcolor{gray!20}\textcolor{red}{71.3}/61.0 & \cellcolor{gray!20}\textcolor{blue}{70.9}/52.4 & \cellcolor{gray!20}\textcolor{blue}{55.7}/59.9 & \cellcolor{gray!20}\textcolor{red}{66.4}/\textcolor{red}{86.8} & \cellcolor{gray!20}\textcolor{red}{84.5}/71.8 & \cellcolor{gray!20}\textcolor{red}{93.6}/74.3 & \cellcolor{gray!20}\textcolor{red}{73.6}/71.3 \\
    
    \midrule
    
    \multirow{4}*{\rotatebox{90}{4-shot}} & \multirow{2}*{TB}
    & M3DM \cite{wang2023multimodal} & 42.4/82.4 & \textcolor{blue}{74.9}/\textcolor{blue}{84.9} & \textcolor{red}{78.7}/\textcolor{blue}{72.1} & \textcolor{blue}{91.4}/\textcolor{red}{96.8} & \textcolor{blue}{70.2}/\textcolor{blue}{80.2} & 53.4/\textcolor{blue}{61.7} & \textcolor{red}{80.2}/\textcolor{red}{81.6} & \textcolor{red}{67.8}/\textcolor{red}{88.6} & 86.6/\textcolor{red}{94.6} & \textcolor{red}{90.6}/\textcolor{red}{92.1} & \textcolor{blue}{73.6}/\textcolor{blue}{83.5} \\
    & & CFM \cite{costanzino2024multimodal} & \textcolor{blue}{43.4}/\textcolor{red}{93.0} & 71.4/\textcolor{red}{85.1} & \textcolor{blue}{78.1}/\textcolor{red}{75.0} & 85.8/\textcolor{blue}{94.1} & \textcolor{red}{72.3}/\textcolor{red}{81.8} & \textcolor{red}{73.0}/\textcolor{red}{75.3} & 50.1/\textcolor{blue}{75.1} & \textcolor{blue}{66.9}/84.3 & \textcolor{red}{95.4}/\textcolor{blue}{93.6} & 79.2/\textcolor{blue}{85.2} & 71.6/\textcolor{red}{84.3} \\
    \cmidrule(lr{0pt}){2-14}
    & TF & \cellcolor{gray!20}CIF (Ours) & \cellcolor{gray!20}\textcolor{red}{48.2}/\textcolor{blue}{87.4} &  \cellcolor{gray!20}\textcolor{red}{81.8}/78.1 & \cellcolor{gray!20}74.6/58.4 & \cellcolor{gray!20}\textcolor{red}{97.9}/88.9 & \cellcolor{gray!20}61.9/68.0 & \cellcolor{gray!20}\textcolor{blue}{69.8}/58.1 & \cellcolor{gray!20}\textcolor{blue}{73.3}/63.9 & \cellcolor{gray!20}61.8/\textcolor{blue}{88.1} & \cellcolor{gray!20}\textcolor{blue}{94.2}/86.0 & \cellcolor{gray!20}\textcolor{blue}{87.4}/78.8 & \cellcolor{gray!20}\textcolor{red}{75.1}/75.6 \\
    \bottomrule
  \end{tabular}
  }
  \label{main_tab}
\end{table*}

\section{Experiments}
\label{exp}

\subsection{Dataset and Evaluation Metrics}

\subsubsection{Dataset}

MVTec 3D-AD \cite{bergmann2021mvtec3d} is a dataset designed for multimodal industrial anomaly detection, containing 10 categories of real-world objects with a total of 4147 samples. Both the training and validation sets consist of only normal samples, while the test set includes various types of anomalies such as scratches, dents, and contamination. Eyecandies \cite{bonfiglioli2022eyecandies} is a synthetic dataset for multimodal anomaly detection, containing 10 categories of candies, with 10000 training images, 1000 validation images, and 4000 test images. Each sample includes RGB images and depth maps, rendered under six different lighting conditions.

\subsubsection{Evaluation Metrics}

We use the area under the receiver operator curve on image-level (I-AUROC) and pixel-level (P-AUROC) to evaluate the detection and segmentation performance of the proposed method respectively. Additionally, to reduce the impact of the anomaly area, we use per-region overlap (PRO) \cite{bergmann2021mvtec} and compute the area under the PRO curve (AUPRO) to evaluate the segmentation performance. AUPRO represents the average overlap of the prediction with each connected component of ground truth.

\subsection{Implementation Details}

When constructing the hypergraph, we set the number of clusters $|\mathcal{E}|$ to 4 for MVTec 3D-AD (for Eyecandies, the number is set to 8). For the training-free hypergraph message passing step, we set the number of message passing steps $L$ to 1 and the retention coefficient $\alpha$ to 0.9. During memory bank construction, we set the sampling rate to 0.1 to ensure an appropriate memory bank size. All experiments were conducted on a single NVIDIA A100 40GB GPU using PyTorch-1.13.1.

\begin{figure*}[t]
\centering
\includegraphics[width=1.0\textwidth]{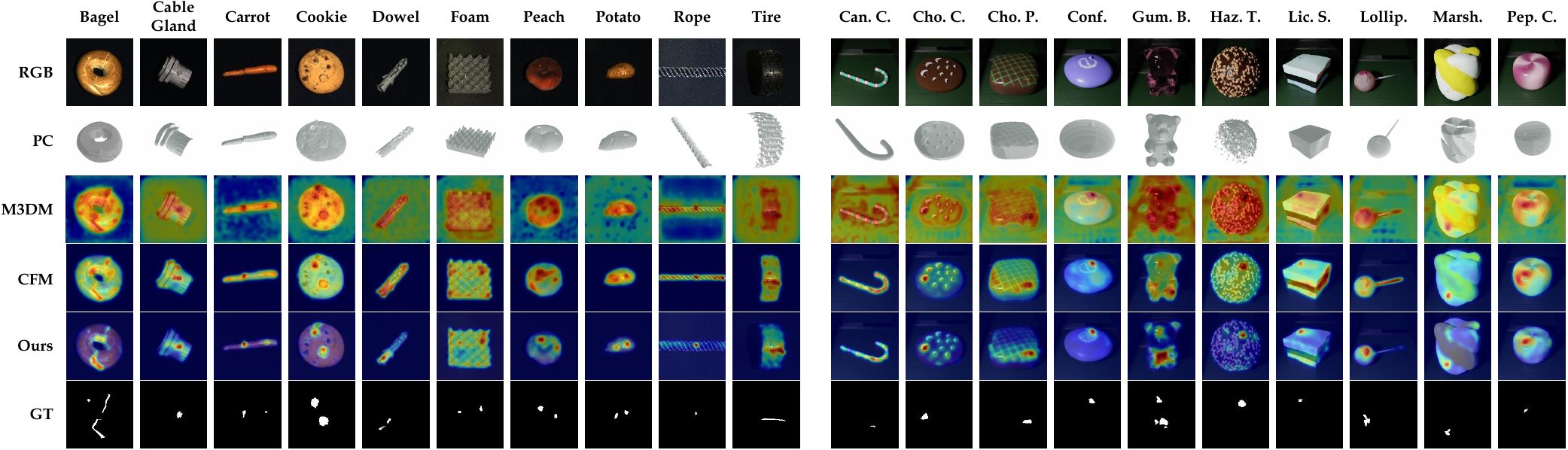}
\caption{Visualization of anomaly scores for each category of MVTec 3D-AD and Eyecandies (in multimodal, 1-shot setting).}
\label{vis}
\end{figure*}

\begin{figure}[t]
\centering
\includegraphics[width=0.9\columnwidth]{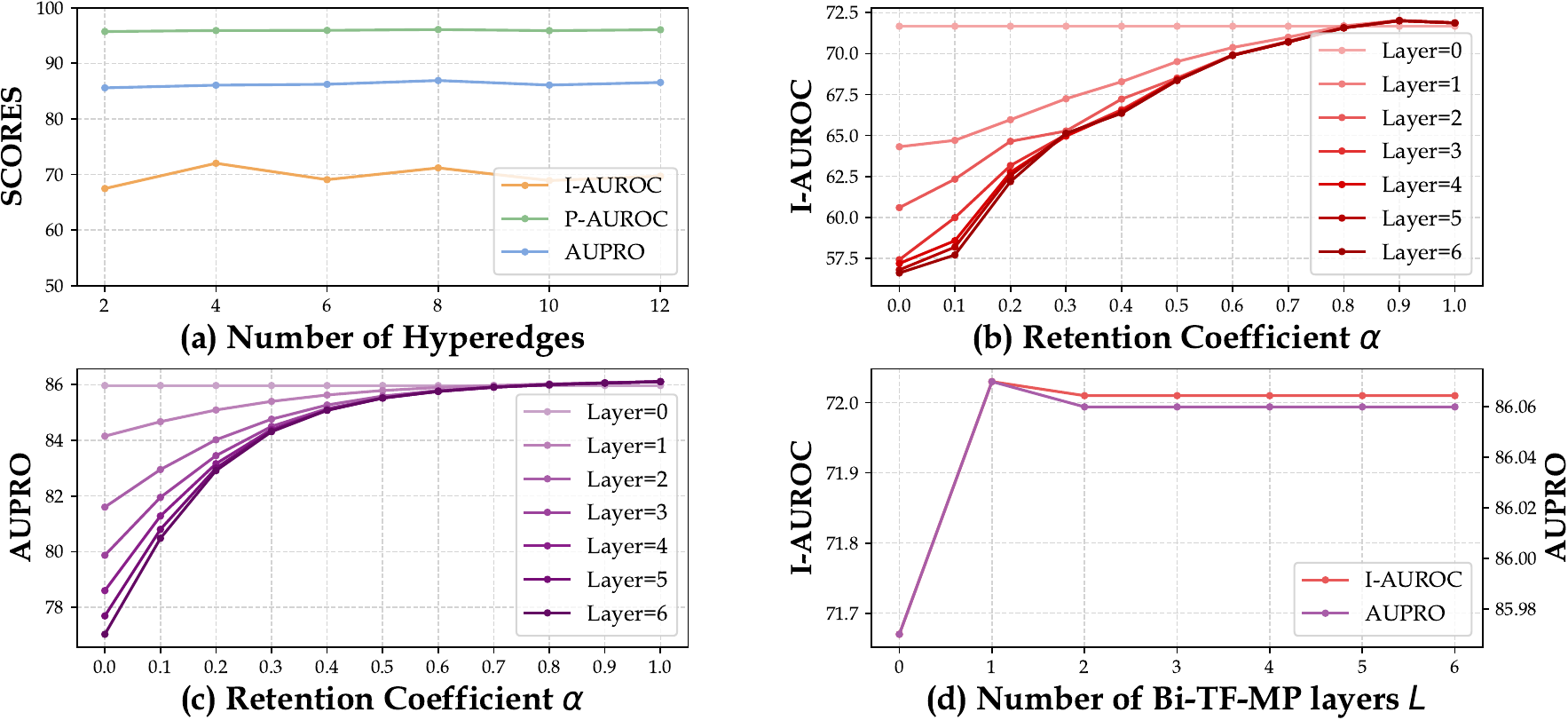}
\caption{Visualization of the impact of parameters in SAHC and Bi-TF-MP on anomaly detection and localization performance.}
\label{vis_abla}
\end{figure}

\subsection{Comparison with SOTAs}

\begin{table}[t]
  \centering
  \caption{Module ablation study of CIF. The best is in \textcolor{red}{red}.}
  \renewcommand\arraystretch{0.8}
  \resizebox{0.4\textwidth}{!}{
  \begin{tabular}{ccccc}
    \toprule
    \multicolumn{3}{c}{Modules} & \multicolumn{2}{c}{Performance} \\
    \cmidrule(lr{0pt}){1-3} \cmidrule(lr{0pt}){4-5}
    SGMS & HGMS & Bi-TF-MP & I-AUROC & AUPRO \\
    \midrule
    & & & 68.6 & 76.1 \\
    $\checkmark$ & & & 71.2 & 85.0 \\
    $\checkmark$ & $\checkmark$ & & 71.7 & 86.0 \\
    $\checkmark$ & $\checkmark$ & $\checkmark$ & \textcolor{red}{72.0} & \textcolor{red}{86.1} \\
    \bottomrule
  \end{tabular}
  }
  \label{ablation}
\end{table}

We compare CIF with several multimodal anomaly detection methods on MVTec 3D-AD and Eyecandies. Table \ref{main_tab} shows anomaly detection performance (I-AUROC scores) and localization performance (AUPRO scores) in few-shot settings. Existing methods are categorized into training-based and training-free approaches. (1) On MVTec 3D-AD, compared with training-free methods, CIF achieves the highest I-AUROC in the 1-shot, 2-shot, and 4-shot settings, outperforming Patchcore+FPFH \cite{horwitz2023back} by 20.2\%, 19.2\%, and 20.6\%, respectively. Compared with training-based methods, although CIF does not achieve the highest I-AUROC, the performance gap with the best method is relatively small: only 2.6\%, 4.5\%, and 3.2\% lower in the 1-shot, 2-shot, and 4-shot settings, respectively. (2) On Eyecandies, we compare CIF with the two best-performing training-based methods on MVTec 3D-AD: M3DM \cite{wang2023multimodal} and CFM \cite{costanzino2024multimodal}. CIF achieves the highest I-AUROC in the 1-shot, 2-shot, and 4-shot settings, outperforming the second-best method by 8.4\%, 7.8\%, and 2.0\%, respectively. (3) On both datasets, using other methods as a baseline, CIF performs better in the 1-shot setting than in the 2-shot and 4-shot settings. This shows that CIF performs better when fewer training samples are available, indicating that the structural information extracted by hypergraphs is effective. (4) On both datasets, CIF does not achieve the highest AUPRO score in any setting, indicating that its ability in anomaly localization is relatively limited. Figure \ref{vis} presents anomaly segmentation visualizations of our method, M3DM, and CFM across all categories on MVTec 3D-AD and Eyecandies (in multimodal, 1-shot setting). "PC" denotes point cloud, and "GT" denotes ground truth. Visually, our method effectively reduces the false positive rate.

\subsection{Ablation Study}

We conduct an ablation study in a multimodal, 1-shot setting on MVTec 3D-AD. To demonstrate our contributions to improving few-shot detection and localization performance, we analyze CIF in Table \ref{ablation} with the following settings: 1) Without SGMS, HGMS, or Bi-TF-MP; 2) With SGMS only; 3) With SGMS and HGMS, but without Bi-TF-MP; 4) With SGMS, HGMS, and Bi-TF-MP (Ours). Comparing row 1 and row 2 in Table \ref{ablation}, we can find that adding SGMS greatly improves the results on both two metrics (I-AUROC by 3.8\% and AUPRO by 11.7\%), indicating that the node features retained through SGMS are more effective. Comparing row 2 and row 3, we observe that HGMS further enhances detection and segmentation performance (I-AUROC by 0.7\% and AUPRO by 1.2\%). Our full model is shown in row 4 in Table \ref{ablation}, and compared with row 3 we have 0.4\% I-AUROC and 0.1\% AUPRO improvement, demonstrating that Bi-TF-MP reduces the distribution gap between test sample nodes and memory bank nodes.

\subsection{Analysis of SAHC Parameter}
\label{sahc_para}

When constructing the hypergraph, the number of hyperedges determines the ability of the hypergraph to capture the structural information of the samples. Too few hyperedges cause different structures to be treated as similar, while too many hyperedges force the same structure to be split apart. We explore the impact of different numbers of hyperedges in SAHC on anomaly detection and localization performance. As shown in Figure \ref{vis_abla} (a), we conducted experiments in a multimodal, 1-shot setting on MVTec 3D-AD, and calculated the mean I-AUROC, P-AUROC, and AUPRO scores across all classes. We found that when the number of hyperedges is 4, I-AUROC reaches its highest; when the number of hyperedges is 8, P-AUROC and AUPRO reach their highest. This shows that fewer hyperedges help improve anomaly detection performance, while more hyperedges benefit anomaly localization performance. We give a class-specific analysis in Appendix.

\subsection{Analysis of Bi-TF-MP Parameter}

During message passing, the retention coefficient $\alpha$ and the number of Bi-TF-MP layers $L$ respectively determine the level of retention of the own information of nodes and the range of information exchange with neighboring nodes. We explore the impact of different $\alpha$ and $L$ values in Bi-TF-MP on anomaly detection and localization performance. We conducted experiments in a multimodal, 1-shot setting on MVTec 3D-AD and calculated the mean I-AUROC and AUPRO scores across all classes. As shown in Figure \ref{vis_abla} (b) and (c), we found that when $\alpha = 0.9$, both I-AUROC and AUPRO reached their highest. This indicates that slight message passing can alleviate the distribution gap between test sample nodes and memory bank nodes. Furthermore, as shown in Figure \ref{vis_abla} (d), when $\alpha = 0.9$, anomaly detection and localization performance were best at $L = 1$, and the performance became stable as $L$ increased. We give a detailed analysis of message passing parameters in Appendix.

\section{Conclusion}
\label{conclusion}

In this paper, we propose a hypergraph-based few-shot multimodal industrial anomaly detection method. We use hypergraphs to extract intra-class structural commonality of samples to guide the construction and search of the memory bank. We introduce semantic-aware hypergraph construction and structure-guided memory sampling to build and compress a memory bank that contains structural information. We then use bidirectional training-free hypergraph message passing to reduce the distribution gap between test samples and the memory bank. Finally, we design hyperedge-guided memory search to reduce the false positive rate in anomaly detection. Our method outperforms the state-of-the-art results on MVTec 3D-AD and Eyecandies in few-shot settings. We hope our work will be helpful for future research.

\section{Acknowledgments}

This work was supported by National Natural Science Foundation of China (No.62576109, 62072112, 62406075), and by National Key Research and Development Program of China (2023YFC3604802).

\bibliography{aaai2026}

\newpage

\section{Appendix}

\subsection{Overview}

The appendix includes the following sections to supplement the theory and experiments in the main text and to highlight the limitations of our method.

\begin{itemize}
\item[—] \textbf{Sec. A} shows the impact of message passing on hypergraph nodes.
\item[—] \textbf{Sec. B} shows the data processing and feature extraction process in our experiment.
\item[—] \textbf{Sec. C} shows the comparison of different hypergraph construction methods.
\item[—] \textbf{Sec. D} shows more visualizations on MVTec 3D-AD and Eyecandies.
\item[—] \textbf{Sec. E} shows more results on MVTec 3D-AD at different numbers of hyperedges.
\item[—] \textbf{Sec. F} shows more ablation study results on MVTec 3D-AD.
\item[—] \textbf{Sec. G} shows the limitations of our method.
\end{itemize}

\section{Impact of message passing on hypergraph nodes}

We use bidirectional training-free hypergraph message passing (Bi-TF-MP) to reduce the distribution gap between test sample node features and memory bank node features. To quantitatively evaluate the impact of Bi-TF-MP on test sample nodes, we introduce the following two metrics:

\begin{itemize}
\item Average Nearest Neighbor Distance (ANND $\downarrow$), measures how close, on average, each test node lies to its closest node in the memory set, with lower values indicating better coverage and similarity.
\item Procrustes Similarity (PCS $\uparrow$), measures how closely the transformed test-node cloud can be optimally rotated and uniformly scaled to match the original, with 1 indicating perfect geometric preservation and 0 indicating complete distortion.
\end{itemize}

The calculation methods for these two metrics are:

\begin{equation}
    \text{ANND}=\frac{1}{N} \sum_{i=1}^{N} \min _{1 \leq j \leq A} d\left(x^{t}_{i}, x^{m}_{j}\right)
\end{equation}

\begin{equation}
\begin{array}{c}
    \text{PCS}=1-\frac{\|\tilde{\mathbf{Y}}-s \tilde{\mathbf{X}} \mathbf{R}\|_{F}^{2}}{\|\tilde{\mathbf{Y}}\|_{F}^{2}}, \\
    \mathbf{R}=\mathbf{U} \mathbf{V}^{\top}, \quad \mathbf{U} \boldsymbol{\Sigma} \mathbf{V}^{\top}=\tilde{\mathbf{Y}}{ }^{\top} \tilde{\mathbf{X}}, \quad s=\frac{\operatorname{tr}(\boldsymbol{\Sigma})}{\|\tilde{\mathbf{X}}\|_{F}^{2}}
\end{array}
\end{equation}

\noindent where $\tilde{\mathbf{X}} = \mathbf{X}^t - \mathbf{1} \bar{\mathbf{x}}^{t\top}$ and $\tilde{\mathbf{Y}} = \mathbf{X}^{t'} - \mathbf{1} \bar{\mathbf{x}}^{t'\top}$. Table \ref{annd_pcs} shows the quantitative impact of Bi-TF-MP on test sample nodes under different values of $\alpha$ and $L$. We observe that as $\alpha$ decreases, less information from the original node is retained, and the distribution gap between the test sample nodes and memory bank nodes becomes smaller after Bi-TF-MP. However, this also introduces a negative effect — the test sample nodes become more distorted, which hinders effective search from the memory bank. Through experiments, we identify a balance between reducing distribution discrepancy and preserving node integrity, with optimal performance achieved at $\alpha = 0.9$ and $L = 1$.

\begin{table*}[t]
  \centering
  \caption{ANND and PCS scores of test sample nodes before and after Bi-TF-MP under different values of $L$ and $\alpha$.}
  \renewcommand\arraystretch{0.8}
  \resizebox{0.8\linewidth}{!}{
  \begin{tabular}{lc|cccccccccc}
    \toprule
    Metrics & Layer & $\alpha=0.9$ & $\alpha=0.8$ & $\alpha=0.7$ & $\alpha=0.6$ & $\alpha=0.5$ & $\alpha=0.4$ & $\alpha=0.3$ & $\alpha=0.2$ & $\alpha=0.1$ & $\alpha=0.0$ \\
    \midrule
    \multicolumn{12}{l}{\cellcolor{gray!20}\textbf{\textit{No Message Passing}}} \\ 
    \midrule
    ANND & - & \multicolumn{10}{c}{53.67} \\
    \midrule
    PCS & - & \multicolumn{10}{c}{100.00} \\
    \midrule
    \multicolumn{12}{l}{\cellcolor{gray!20}\textbf{\textit{Bi-TF-MP}}} \\
    \midrule
    \multirow{6}*{{ANND}}
    & 1 & 52.18 & 50.75 & 49.36 & 48.02 & 46.74 & 45.53 & 44.40 & 43.33 & 42.35 & 41.46 \\
    & 2 & 52.12 & 50.51 & 48.86 & 47.18 & 45.53 & 43.94 & 42.47 & 41.17 & 40.10 & 39.34 \\
    & 3 & 52.12 & 50.49 & 48.79 & 47.04 & 45.29 & 43.60 & 42.05 & 40.77 & 39.88 & 39.57 \\
    & 4 & 52.12 & 50.49 & 48.78 & 47.02 & 45.24 & 43.52 & 41.97 & 40.74 & 40.06 & 40.09 \\
    & 5 & 52.12 & 50.49 & 48.78 & 47.01 & 45.23 & 43.51 & 41.97 & 40.80 & 40.27 & 40.39 \\
    & 6 & 52.12 & 50.49 & 48.78 & 47.01 & 45.23 & 43.51 & 41.98 & 40.85 & 40.45 & 40.53 \\
    \midrule
    \multirow{6}*{{PCS}}
    & 1 & 99.95 & 99.80 & 99.50 & 99.03 & 98.35 & 97.41 & 96.15 & 94.47 & 92.29 & 89.58 \\
    & 2 & 99.95 & 99.77 & 99.39 & 98.71 & 97.59 & 95.77 & 92.93 & 88.45 & 81.50 & 70.92 \\
    & 3 & 99.95 & 99.77 & 99.38 & 98.67 & 97.44 & 95.32 & 91.69 & 85.34 & 74.16 & 55.09 \\
    & 4 & 99.95 & 99.76 & 99.38 & 98.66 & 97.41 & 95.22 & 91.30 & 83.99 & 70.07 & 44.51 \\
    & 5 & 99.95 & 99.77 & 99.38 & 98.66 & 97.41 & 95.22 & 91.22 & 83.54 & 68.00 & 37.71 \\
    & 6 & 99.95 & 99.76 & 99.38 & 98.67 & 97.42 & 95.23 & 91.24 & 83.45 & 66.93 & 33.28 \\
    \bottomrule
  \end{tabular}
  }
  \label{annd_pcs}
\end{table*}

\begin{table*}[t]
  \centering
  \caption{I‑AUROC/AUPRO scores for anomaly detection and localization of all categories of MVTec 3D‑AD in 1‑shot setting under different numbers of hyperedges.}
  \renewcommand\arraystretch{0.8}
  \resizebox{1.0\linewidth}{!}{
  \begin{tabular}{cccccccccccc}
    \toprule
    Hyperedge Num & Bagel & Cable Gland & Carrot & Cookie & Dowel & Foam & Peach & Potato & Rope & Tire & Mean \\
    \midrule
    2 & 59.1/84.2 & 69.5/79.6 & 71.5/95.6 & 86.3/83.8 & 63.2/82.1 & 62.8/68.7 & 69.9/93.4 & 60.2/94.3 & 83.4/87.8 & 48.7/86.5 & 67.5/85.6 \\
    4 & 78.9/85.2 & 68.6/79.5 & 72.1/95.6 & 81.2/86.2 & 62.8/82.0 & 72.8/70.3 & 84.0/93.3 & 60.3/94.2 & 83.5/88.1 & 56.0/86.3 & 72.0/86.1 \\
    6 & 76.7/85.5 & 67.3/80.0 & 69.8/95.6 & 69.9/86.1 & 63.6/82.1 & 74.7/72.1 & 75.3/93.4 & 57.3/94.3 & 84.5/88.0 & 51.5/85.1 & 69.1/86.2 \\
    8 & 79.6/87.9 & 69.3/80.2 & 70.5/95.6 & 81.1/88.9 & 64.0/82.0 & 72.7/71.5 & 82.1/94.1 & 55.7/94.3 & 84.6/87.9 & 52.3/86.6 & 71.2/86.9 \\
    10 & 77.3/86.5 & 70.3/80.1 & 69.9/95.6 & 80.4/87.0 & 62.5/82.0 & 64.4/69.5 & 82.8/93.8 & 58.9/94.3 & 82.5/88.1 & 40.1/84.0 & 68.9/86.1 \\
    12 & 75.6/86.5 & 69.6/80.0 & 70.9/95.7 & 76.6/87.9 & 59.9/82.0 & 73.1/71.0 & 84.2/94.2 & 58.9/94.0 & 82.7/88.1 & 45.4/86.3 & 69.7/86.6 \\
    \bottomrule
  \end{tabular}
  }
  \label{hyperedge_more}
\end{table*}

\begin{table*}[t]
  \centering
  \caption{I‑AUROC/AUPRO scores for anomaly detection and localization of all categories of MVTec 3D‑AD in 1‑shot setting.}
  \renewcommand\arraystretch{0.8}
  \resizebox{1.0\linewidth}{!}{
  \begin{tabular}{cccccccccccccc}
    \toprule
    \multicolumn{3}{c}{Modules} & \multirow{2}{*}{Bagel} & \multirow{2}{*}{Cable Gland} & \multirow{2}{*}{Carrot} & \multirow{2}{*}{Cookie} & \multirow{2}{*}{Dowel} & \multirow{2}{*}{Foam} & \multirow{2}{*}{Peach} & \multirow{2}{*}{Potato} & \multirow{2}{*}{Rope} & \multirow{2}{*}{Tire} & \multirow{2}{*}{Mean} \\
    \cmidrule(lr{0pt}){1-3}
    SGMS & HGMS & Bi-TF-MP \\
    \midrule
    & & & 72.1/82.6 & 67.7/79.3 & 68.5/90.7 & 79.5/85.3 & 55.1/78.0 & 67.1/53.7 & 85.5/90.5 & 55.9/55.9 & 86.2/70.8 & 48.2/74.0 & 68.6/76.1 \\
    $\checkmark$ & & & 77.2/81.2 & 69.5/79.1 & 71.0/95.6 & 80.1/81.8 & 62.9/81.2 & 70.4/73.7 & 84.4/92.0 & 59.0/94.1 & 83.2/85.6 & 54.6/85.7 & 71.2/85.0 \\
    $\checkmark$ & $\checkmark$ & & 78.1/84.3 & 70.1/79.5 & 71.0/95.6 & 80.1/85.7 & 63.3/82.0 & 71.3/82.0 & 84.5/71.3 & 60.5/92.9 & 82.8/94.1 & 55.1/88.0 & 71.7/86.0 \\
    $\checkmark$ & $\checkmark$ & $\checkmark$ & 78.9/85.2 & 68.7/79.5 & 72.2/95.6 & 81.2/86.2 & 62.9/82.0 & 72.8/70.3 & 83.9/93.3 & 60.4/94.2 & 83.5/88.1 & 55.8/86.4 & 72.0/86.1 \\
    \bottomrule
  \end{tabular}
  }
  \label{ablation_more}
\end{table*}

\begin{figure*}[t]
    \centering
    \includegraphics[width=0.9\textwidth]{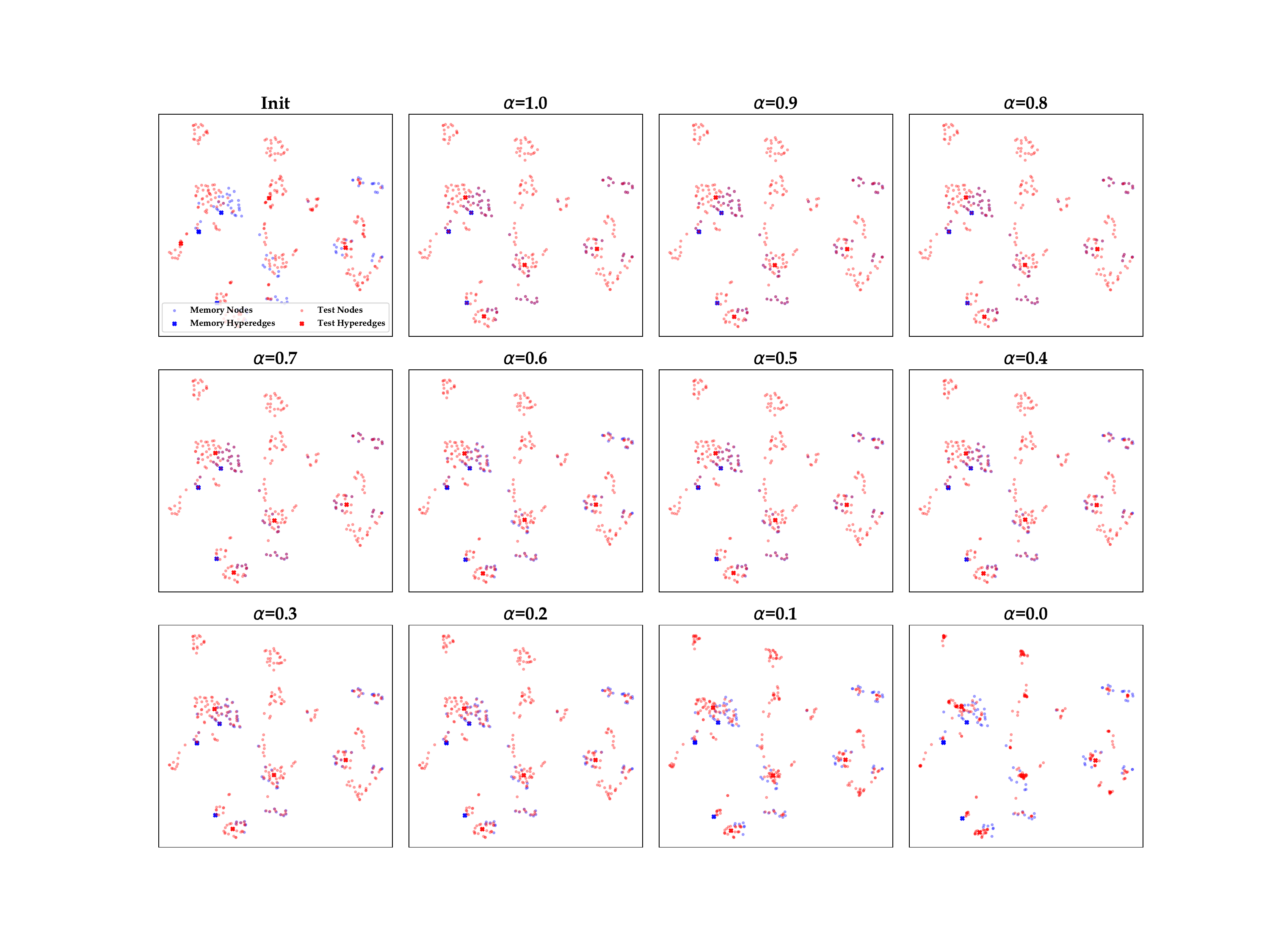}
    \caption{Distribution of test sample features and memory bank features before and after bidirectional training-free hypergraph message passing.}
    \label{mp_vis}
\end{figure*}

\begin{figure*}[t]
    \centering
    \includegraphics[width=0.8\textwidth]{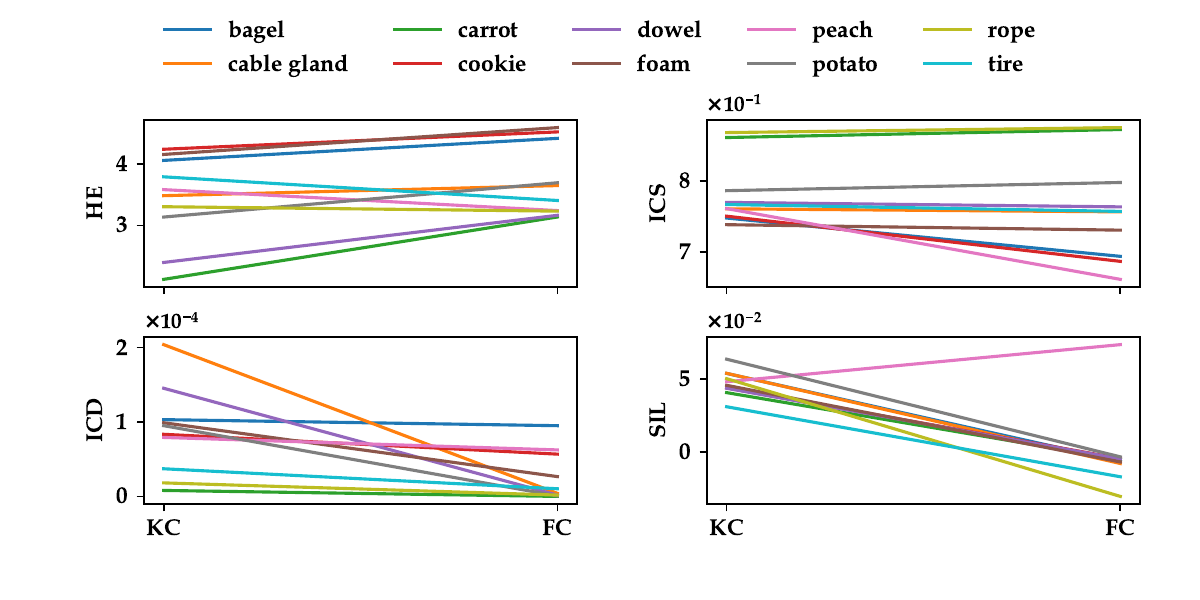}
    \caption{Comparing hypergraph quality metrics on MVTec 3D-AD (FCM represent Fuzzy C-Means).}
    \label{hyperscores}
\end{figure*}

To visually show the impact of message passing on node features, we apply UMAP \cite{mcinnes2018umap} to reduce the dimension of both the test sample node features and memory bank node features, and visualize the node features with $\alpha$ varying from 1 to 0 ($L=1$), as shown in Figure \ref{mp_vis}. A larger $\alpha$ indicates more retention of the own information of the node. We find that a larger $\alpha$ effectively mitigates the distribution gap between test sample nodes and memory bank nodes without distorting their original distributions. In contrast, a smaller $\alpha$ causes nodes to overly absorb information from neighbor nodes, leading to the distortion of their original feature distributions.

\section{Data Processing and Feature Extraction}

For RGB images, we resize them to $224 \times 224$ using bilinear interpolation and normalize the resulting tensors. Because DINO \cite{caron2021emerging} pretraining method effectively fuses local and global information, we extract patch features with a Vision Transformer (ViT) \cite{dosovitskiy2020image} model trained using DINO. Specifically, we adopt a ViT-B/8 architecture and employ the model that was pretrained on ImageNet \cite{deng2009imagenet} to obtain more stable patch representations. The model recieves an input image of size $224 \times 224$ and divides it into image blocks of size $8 \times 8$, and generating $28 \times 28$ non-overlapping patches, each patch is encoded as a feature embedding with 768 dimensions. These patch features exhibit strong discriminative power and are well suited for downstream unsupervised anomaly detection tasks.

For 3D point clouds, we resize the point cloud tensors to $224 \times 224$ with nearest neighbor interpolation method, ensuring that their spatial structure aligns with the image input resolution. Refer to M3DM \cite{wang2023multimodal}, we use a Point Transformer \cite{zhao2021point,pang2022masked} pretrained on ShapeNet dataset \cite{chang2015shapenet} to extract patch-level point cloud features. After removing zero points, we apply farthest point sampling (FPS) \cite{qi2017pointnet++} to generate $M = 1024$ groups, each with $S = 128$ points. These point groups are processed by a local encoder and then passed to a global attention module composed of multiple Transformer blocks. We concatenate the ${3, 7, 11}$ layer output to obtain point cloud features with 1152 dimensions. To match the size of the RGB image features, we employ adaptive average pooling to resize the point cloud feature maps to $28 \times 28$.

To mitigate background interference during detection, we follow AST \cite{rudolph2023asymmetric} and use 3D point clouds to extract foreground masks. We first obtain the depth channel of the point cloud, and fill missing depth values with $3$ iterations of neighborhood mean interpolation after normalization. The background is estimated from the corner depth values, the distances between all other areas and this background are computed, and extract the foreground mask by setting the threshold to $7 \times 10^{-3}$.

\section{Comparison of Hypergraph Construction Methods}

Previous methods \cite{han2023vision} use Fuzzy C-Means \cite{bezdek1984fcm} to construct hypergraphs for images, but Fuzzy C-Means performs poorly on single-semantic industrial images. We perform both quantitative and qualitative comparisons between our SAHC and Fuzzy C-Means. According to HgVT \cite{fixelle2025hypergraph}, we use the following four metrics to evaluate the quality of the constructed hypergraph ($\uparrow$ means the higher the better, $\downarrow$ means the lower the better):

\begin{itemize}
\item Hyperedge Entropy (HE $\downarrow$), measures the concentration of vertex features within each hyperedge.
\item Intra-Cluster Similarity (ICS $\uparrow$), measures the cohesion of vertex features within each hyperedge.
\item Inter-Cluster Distance (ICD $\uparrow$), measures how distinct different hyperedges are from one another.
\item Silhouette Score (SIL $\uparrow$), measures how well each vertex is clustered with respect to its assigned hyperedge and nearby hyperedges.
\end{itemize}

We use Fuzzy C-Means to construct hypergraphs only on foreground nodes and compare the constructed hypergraphs with those constructed using our SAHC. As shown in Figure \ref{hyperscores}, we calculated four scores for the hypergraphs constructed using SAHC and Fuzzy C-Means across all categories in MVTec 3D-AD. Overall, the hypergraphs constructed using SAHC have a lower HE and higher ICS, ICD, and SIL.

\begin{figure*}[h]
    \centering
    \includegraphics[width=0.9\textwidth]{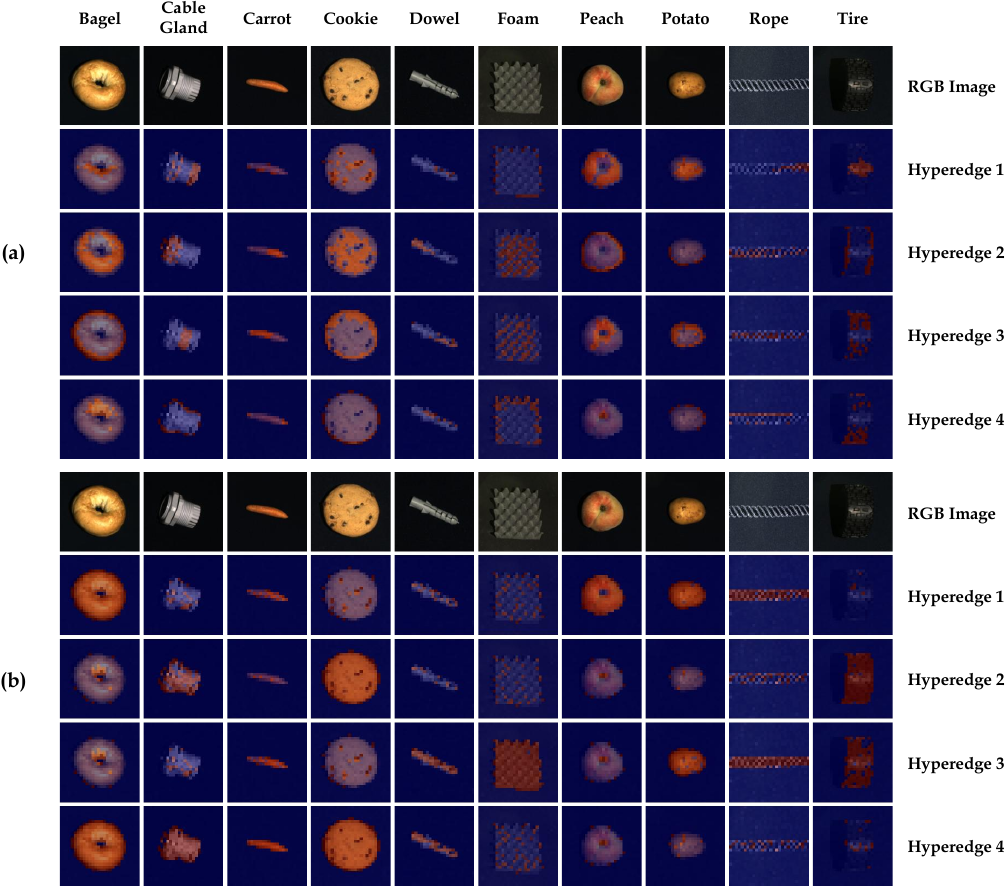}
    \caption{Visualization of Hyperedges on MVTec 3D-AD.}
    \label{hypergraph_vis}
\end{figure*}

We also visualize the hyperedge distribution of the hypergraphs constructed using SAHC and Fuzzy C-Means. As shown in Figure \ref{hypergraph_vis}, (a) is the visualization of the hyperedges in the hypergraph constructed using SAHC, and (b) is the visualization of the hyperedges in the hypergraph constructed using Fuzzy C-Means. Visually, our SAHC constructs a more evenly distributed hypergraph, better capturing the structure of the samples. We find that Fuzzy C-Means performs better on multi-semantic images in open-world scenarios. However, on single-semantic industrial images, even when Fuzzy C-Means is applied only to foreground patches during comparison, it still fails to effectively distinguish highly similar patch features. This indicates that for single-semantic industrial images, it is still necessary to use a hard clustering algorithm to first determine the hyperedge centers and then assign nodes based on similarity.

\section{More Visualizations}

\begin{figure*}[h]
    \centering
    \includegraphics[width=1.0\textwidth]{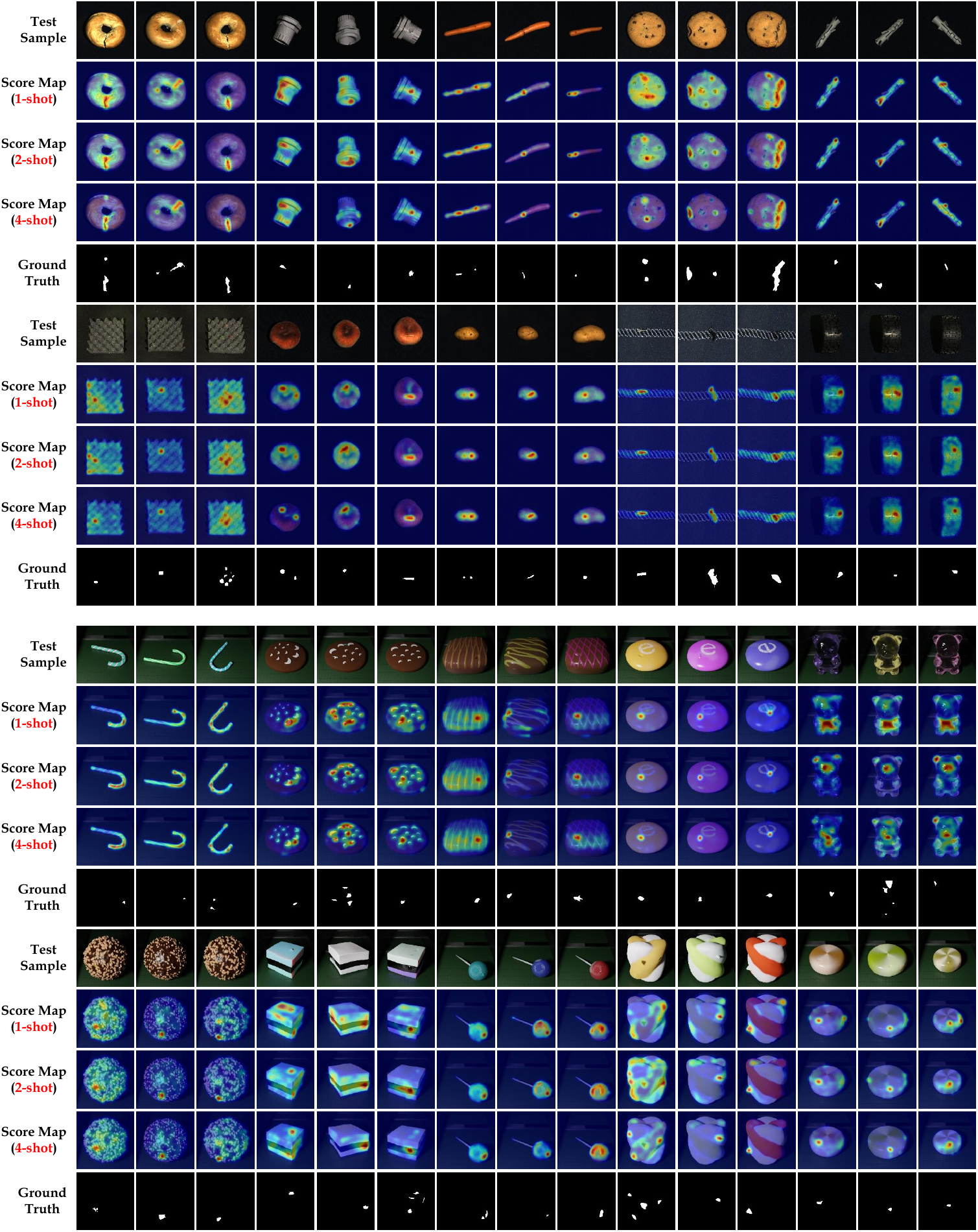}
    \caption{Visualization of anomaly segmentation results for each category of our method on MVTec 3D-AD and Eyecandies.}
    \label{vis_more_all}
\end{figure*}

We supplement more segmentation results for each category on MVTec 3D-AD and Eyecandies (including 1-shot, 2-shot, and 4-shot results) in Figure \ref{vis_more_all}.

\section{More Results at Different Numbers of Hyperedges}

Table \ref{hyperedge_more} presents the anomaly detection (I-AUROC) and anomaly localization (AUPRO) results for all categories in MVTec 3D-AD under different numbers of hyperedges, as discussed in Section 4.5.

Although the visualization of the impact of parameters in SAHC (see Section 4.5 for details) does not show a significant impact of the number of hyperedges on anomaly detection and localization performance, we can observe from Table \ref{hyperedge_more} that different numbers of hyperedges have a significant effect on the anomaly detection and localization performance for different class samples. For example, in Section 4.5, we find that the best performance in anomaly detection and localization occurs when the number of hyperedges is 4 or 8. However, in Table \ref{hyperedge_more}, the Cable Gland, Cookie, Foam, and Peach classes achieve the best anomaly detection performance with 10, 2, 6, and 12 hyperedges, respectively. This shows that the structural complexity of different classes within the same dataset varies, and different numbers of hyperedges are beneficial for capturing structural commonalities of varying levels of complexity.

\section{More Ablation Study Results}

Table \ref{ablation_more} presents the anomaly detection (I-AUROC) and anomaly localization (AUPRO) results for all categories in MVTec 3D-AD in Section 4.4.

\section{Limitations}

Our method has the following limitations, which we aim to address in future work:

\begin{itemize}
\item In full-shot setting, as the number of nodes stored in the memory bank increases, the computational cost of updating hyperedge features in our SAHC grows accordingly, with a complexity of $O(N*|\mathcal{E}|*dim)$. This makes our method currently unsuitable for scenarios involving large-scale training data.
\item Our SAHC currently constructs hypergraphs only on RGB images, while hypergraphs constructed on 3D point clouds fail to capture effective structural information. This indicates that our methods are not well-suited for point-based features, and consequently, our approach is currently not applicable to pure 3D anomaly detection tasks.
\end{itemize}

\end{document}